\documentclass[runningheads]{llncs}
\usepackage[T1]{fontenc}
\usepackage[utf8]{inputenc}
\usepackage{newunicodechar}
\newunicodechar{✉}{\textemail}
\usepackage{placeins}

\usepackage{graphicx}
\usepackage{multirow}
\usepackage{cite}
\usepackage{amsmath,amssymb,amsfonts}
\usepackage{algorithmic}
\usepackage{textcomp}
\usepackage{xcolor}
\usepackage{subcaption}  
\usepackage{makecell}
\usepackage{graphicx}
\usepackage{subcaption}  
\begin{document}

\title{Enhancing YOLOv11n for Reliable Child Detection in Noisy Surveillance Footage}
\titlerunning{Enhancing YOLOv11n for Child Detection}

\author{
Khanh Linh Tran
\and
Minh Nguyen Dang
\and
Thien Nguyen Trong
\and
Hung Nguyen Quoc
\and
Linh Nguyen Kieu\thanks{Corresponding author: linhnk@ptit.edu.vn} 
}

\authorrunning{Khanh Linh Tran et al.}
\institute{Posts and Telecommunications Institute of Technology, Hanoi, Vietnam}

\maketitle              
\begin{abstract}
This paper presents a practical and lightweight solution for enhancing child detection in low-quality surveillance footage, a critical component in real-world missing child alert and daycare monitoring systems. Building upon the efficient YOLOv11n architecture, we propose a deployment-ready pipeline that improves detection under challenging conditions including occlusion, small object size, low resolution, motion blur, and poor lighting, common in existing CCTV infrastructures. Our approach introduces a domain-specific augmentation strategy that synthesizes realistic child placements using spatial perturbations (e.g., partial visibility, truncation, and overlaps) combined with photometric degradations (e.g., lighting variation and noise). To improve recall of small and partially occluded instances, we integrate Slicing Aided Hyper Inference (SAHI) at inference time. All components are trained and evaluated on a filtered, child-only subset of the Roboflow Daycare dataset. Compared to the baseline YOLOv11n, our enhanced system achieves a mAP@0.5 of 0.967 and mAP@0.5:0.95 of 0.783, yielding absolute improvements of 0.7\% and 2.3\% respectively, without architectural changes. Importantly, the entire pipeline maintains compatibility with low-power edge devices and supports real-time performance, making it particularly well-suited for low-cost or resource-constrained industrial surveillance deployments. The example augmented dataset and the source code used to generate it are available at: \url{https://github.com/html-ptit/Data-Augmentation-YOLOv11n-child-detection}.

\end{abstract}

\section{Introduction}

Reliable child detection in surveillance footage is essential for public safety systems in daycares, schools, and playgrounds, enabling real-time monitoring, anomaly detection, and rapid response in missing child scenarios, an integral part of smart city safety infrastructure~\cite{inkaya2020yolov3, moussa2023children}. However, real-world detection remains challenging due to children’s small size, frequent occlusions, rapid movements, and the low quality of typical CCTV footage, which often suffers from low resolution, motion blur, compression artifacts, and poor lighting~\cite{khanam2024yolov11, nguyen2020smallobject}. These factors severely hinder the performance of modern YOLO-based detectors~\cite{chen2020ship, hu2011superres}.

The YOLO (You Only Look Once) family of object detectors has been widely adopted across industries due to its high speed and real-time inference capabilities. Recent variants such as YOLOv8, YOLOv10, and YOLOv11 have made incremental improvements in speed and accuracy, offering lightweight models like YOLOv11n that are well-suited for deployment on edge devices in industrial environments~\cite{terven2023review, sapkota2024yolov10, jegham2024benchmark}. Nonetheless, these models are not explicitly optimized for detecting small or partially visible children in surveillance-style scenes, where performance degradation remains a major bottleneck~\cite{rasheed2024optimization}.

In this paper, we propose a practical and deployment-friendly enhancement of YOLOv11n for child detection in noisy surveillance footage. Our approach fine-tunes YOLOv11n on a curated, single-class dataset derived from the Roboflow Daycare dataset~\cite{roboflow_daycare}, containing only child instances. We introduce a domain-specific data augmentation pipeline that simulates real-world surveillance distortions, such as object overlap, edge truncation, camera noise, low lighting, and unusual angles, to improve model robustness. Furthermore, we integrate Slicing Aided Hyper Inference (SAHI)~\cite{akyon2022sahi} as a lightweight inference-time strategy to increase recall for small and partially occluded children, a frequent occurrence in surveillance footage.

\noindent\textit{The main contributions of this paper are as follows:}
\begin{enumerate}
  \item A curated child-only dataset from the Roboflow Daycare dataset, filtered and adapted for surveillance-specific object detection.
  \item A realistic augmentation pipeline that simulates common surveillance artifacts to improve generalization under operational conditions.
  \item An enhancement of YOLOv11n with SAHI for improved detection of small children while preserving deployment compatibility.
  \item An evaluation under conditions representative of real-world deployments, demonstrating improved accuracy and robustness over the baseline.
\end{enumerate}

\section{Related Work}

Human detection in surveillance has long been a focus in computer vision for safety and behavioral monitoring. The YOLO (You Only Look Once) family of single-stage detectors is widely recognized for real-time performance. Since YOLOv1~\cite{redmon2016you}, later versions such as YOLOv4 and YOLOv7~\cite{wang2023yolov7} improved detection through architectural and training refinements. Recent variants, including YOLOv8, YOLOv10, and YOLOv11, further optimize modular design, decoupled heads, and label assignment for better accuracy–efficiency trade-offs~\cite{sapkota2024yolov10, jegham2024benchmark}. YOLOv11n, the lightweight model used here, suits edge devices but still struggles with small or occluded targets common in surveillance.

Small object detection remains difficult. SSD~\cite{liu2016ssd} leverages multi-scale features, while attention- or transformer-based approaches~\cite{li2023transformersmall, zhang2023sr} enhance accuracy at the cost of latency. Akyon et al.~\cite{akyon2022sahi} introduced SAHI, which improves small object detection via patch-based inference without retraining, complementing YOLO architectures.

Child detection, in contrast, is underrepresented in datasets such as COCO~\cite{lin2014microsoft} and CrowdHuman~\cite{shao2018crowdhuman}. Due to children’s smaller size and distinct posture, existing models~\cite{moussa2023children, inkaya2020yolov3} show reduced accuracy under occlusion and poor lighting. To address this, we fine-tune YOLOv11n on a child-only subset of the Roboflow Daycare dataset for single-class detection.

Robustness against surveillance noise has been pursued through augmentation techniques such as motion blur, lighting shifts, and noise injection~\cite{zhao2021augmenting, cheng2022lightweight, dai2021generalizable}, often supported by Albumentations~\cite{buslaev2020albumentations}. However, these are typically random rather than scene-aware. Our work extends this by introducing a domain-specific augmentation pipeline that composites segmented children into real scenes with occlusion, truncation, and lighting distortions, explicitly modeling low-end CCTV artifacts to enhance robustness and realism.

\section{Methodolody}

\subsection{Overview}
Our pipeline aims to improve the robustness of human detection, particularly for children, in noisy surveillance environments. The core of our approach involves fine-tuning the YOLOv11n object detector on a surveillance-style dataset enhanced through custom augmentations. To further improve detection of small and partially occluded subjects, we integrate SAHI (Slicing Aided Hyper Inference) during inference. This section details our dataset preparation, augmentation pipeline, model fine-tuning strategy, and integration with SAHI.

\subsection{Dataset Preparation}
We use the publicly available Daycare dataset from Roboflow Universe \cite{roboflow_daycare}, which consists of surveillance-like imagery from daycare environments, containing labeled instances of both children and adults. The dataset aligns well with our objectives, featuring realistic conditions such as crowded indoor environments, fluctuating lighting, and frequent occlusions. For this work, we exclusively retain annotations corresponding to children and discard adult samples to tailor the model for child-specific detection.

Before training, we conducted a preprocessing and cleaning phase to ensure data quality. Together, these augmentations strengthen model robustness by exposing the detector to a diverse range of cases where bounding boxes were misaligned or missing. We also unified the label schema by mapping all remaining instances to a single class: \texttt{child}. 

The final dataset is split into three subsets: training, validation, and testing, as shown in Table~\ref{tab:dataset_split}.

\begin{table}[htbp]
\caption{Dataset Split Statistics (Child Instances Only)}
\centering
\begin{tabular}{|c|c|c|c|}
\hline
\textbf{Subset} & \textbf{Images} & \textbf{Child Instances} & \textbf{Percentage} \\
\hline
Training   & 2289 & 15518 & 69\% \\
Validation & 447  & 3067  & 14\% \\
Testing    & 564  & 3755  & 17\% \\
\hline
\textbf{Total}     & \textbf{3300} & \textbf{22340} & \textbf{100\%} \\
\hline
\end{tabular}
\label{tab:dataset_split}
\end{table}

All annotations were converted into YOLO-compatible format, where each bounding box is defined by a class ID followed by normalized center coordinates and width-height dimensions. Since we model only one class (\texttt{child}), the class ID is consistently set to 0 throughout the dataset.

\subsection{Data Augmentation Strategy}
\label{subsection:DataAugmentation}

To improve detection robustness under challenging surveillance conditions, we design a two-part augmentation strategy that introduces synthetic child figures and simulates visual noise. The motivation behind this approach is to mimic real-world factors like occlusion, poor lighting, and truncation, which are commonly found in surveillance footage. The pipeline consists of two main augmentation types: (1) object-level augmentations through synthetic instance compositing, and (2) image-level degradation augmentations. Each augmentation is applied with a defined probability during training.

\subsubsection{Synthetic Child Instance Compositing}
This augmentation type aims to increase the number of child instances in training images while mimicking spatial and contextual conditions typical in real surveillance scenes. This process consists of several steps:

\begin{itemize}
    \item \textbf{Segmentation and Synthetic Instance Creation}: Child figures are first segmented from existing labeled images to create a repository of reusable child cutouts. We use bounding box masks and basic background removal techniques to isolate clean silhouettes of children.

    \item \textbf{Scaling and Rotation}: To ensure natural scaling relative to existing objects, each pasted child is scaled based on the distribution of bounding box areas already present in the target image. Rotation is applied randomly in a small range to simulate different postures and perspectives, which is common in real-world scenes.

    \item \textbf{Positioning Strategies}: For each image, one of the following positioning strategies is selected to place new child instances. Figure~\ref{fig:synthetic_compositing} illustrates examples of our synthetic child instance compositing strategy, showcasing the original image alongside augmented variants where synthetic children are added overlap on existed child, near the image edge, and in the center area:
    \begin{itemize}
        \item \textit{Occlusion Simulation}: Child cutouts are pasted so their bounding boxes overlap with existing ones, but the new instance’s center does not coincide with that of any existing bounding box. This ensures partial occlusion without full redundancy.
        \item \textit{Edge Truncation}: Children are positioned so that a portion of their bounding box falls outside the visible image frame. A constraint is enforced to ensure that a minimum area of the added bounding box (e.g., at least 50\%) remains inside the image.
        \item \textit{Neutral Placement}: Children are placed randomly within central areas of the image, simulating less obstructed viewpoints.
    \end{itemize}

    \item \textbf{Blending and Scene Adaptation}: To make synthetic children appear naturally within the scene, we apply alpha blending that accounts for local image brightness and noise. Brightness is estimated using a simple grayscale mask, and local noise level is assessed using the Laplacian variance. These metrics guide the adjustment of pasted child figures for better photometric consistency with their target region.
\begin{figure*}[h]
  \centering
  \begin{subfigure}{0.24\linewidth}
    \includegraphics[width=\linewidth]{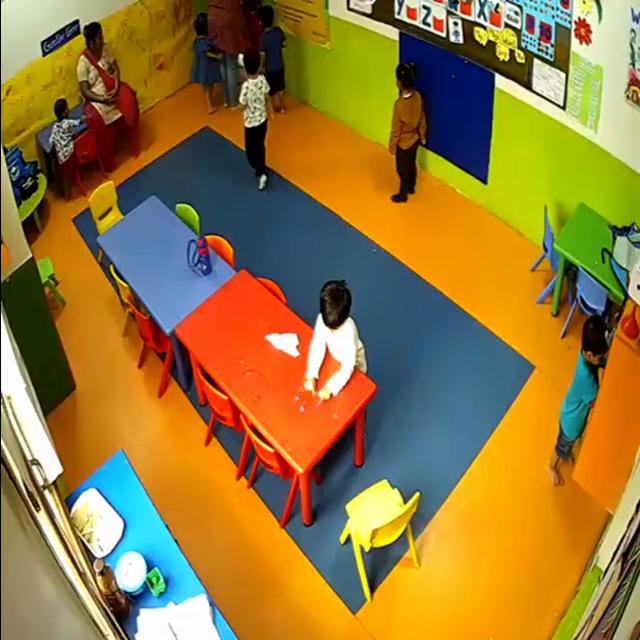}
    \caption{Original image from daycare training dataset}
  \end{subfigure}
  \begin{subfigure}{0.24\linewidth}
    \includegraphics[width=\linewidth]{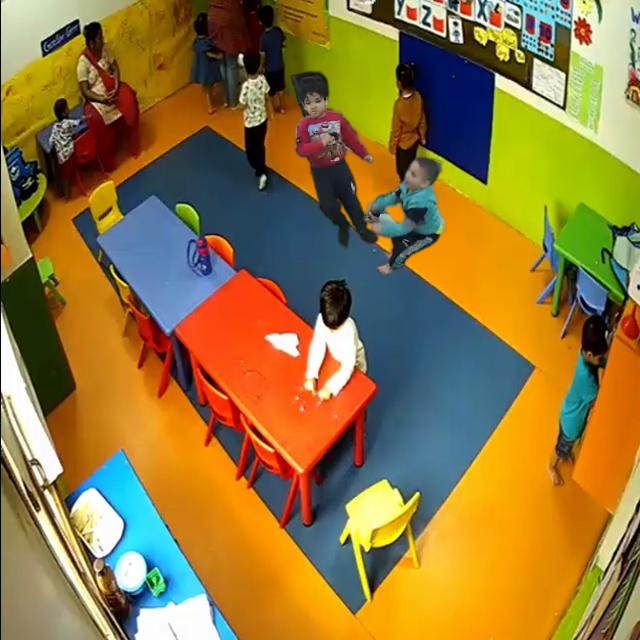}
    \caption{Synthetic child added overlap existed child}
  \end{subfigure}
  \begin{subfigure}{0.24\linewidth}
    \includegraphics[width=\linewidth]{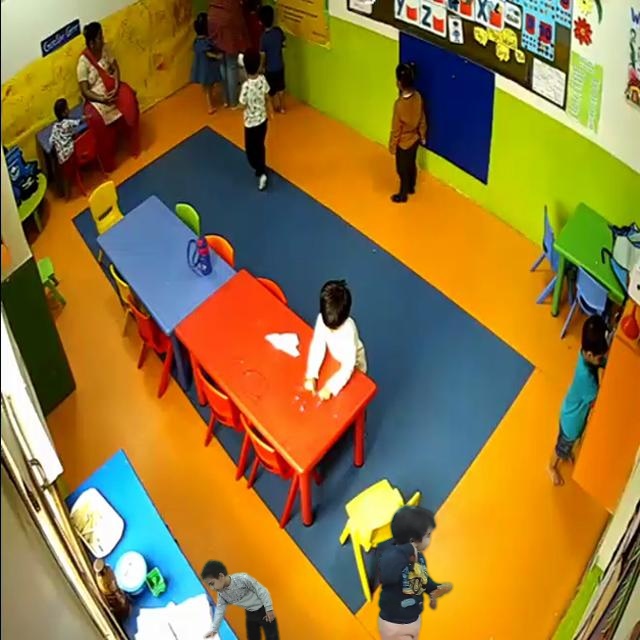}
    \caption{Synthetic child added near frame edge}
  \end{subfigure}
  \begin{subfigure}{0.24\linewidth}
    
    \includegraphics[width=\linewidth]{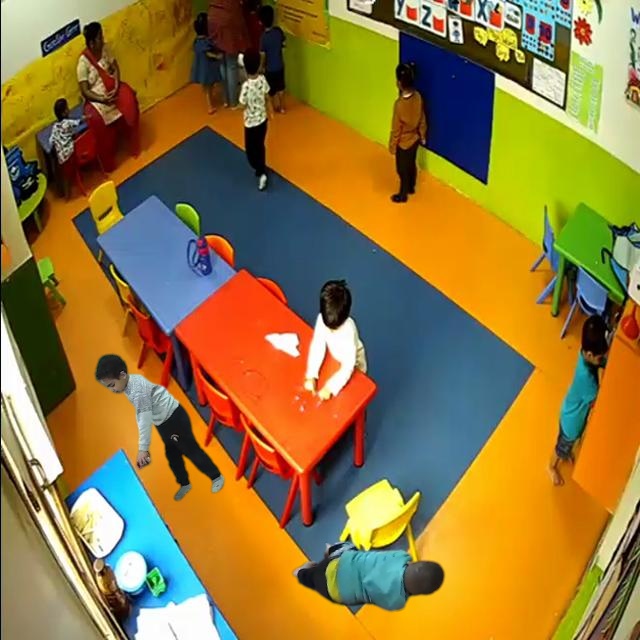}
    \caption{Synthetic child added in center \\ area}
  \end{subfigure}
  \caption{Synthetic Child Instance Compositing examples with different positioning strategies.}
  \label{fig:synthetic_compositing}
\end{figure*}
\end{itemize}

\vspace{-3em}
\subsubsection{Image-Level Degradations}

To further simulate surveillance artifacts, we apply degradations that mimic common noise patterns and lighting variations observed in real-world footage.

\begin{itemize}
    \item \textbf{Noise Injection}: Each image has a chance to be augmented with one or two types of noise artifacts. Available options include:
    \begin{itemize}
        \item \textit{Additive Stripe Noise}: Simulates compression or analog video artifacts common in older or low-quality surveillance systems.
        \item \textit{Spatial Filtering}: Introduces blur or pixel-level corruption to mimic motion blur and camera shake.
    \end{itemize}

    \item \textbf{Lighting Adjustments}: One or two lighting-based augmentations are applied per image. These include:
    \begin{itemize}
        \item \textit{Histogram Equalization}: Enhances contrast in low-light images.
        \item \textit{Contrast Stretching}: Expands the intensity range to improve visibility under uneven lighting.
        \item \textit{Color Grading}: Alters the color tone to simulate tinted or poorly white-balanced footage.
    \end{itemize}
\end{itemize}

\begin{figure}[!htbp]
  \centering
  \includegraphics[width=0.9\linewidth]{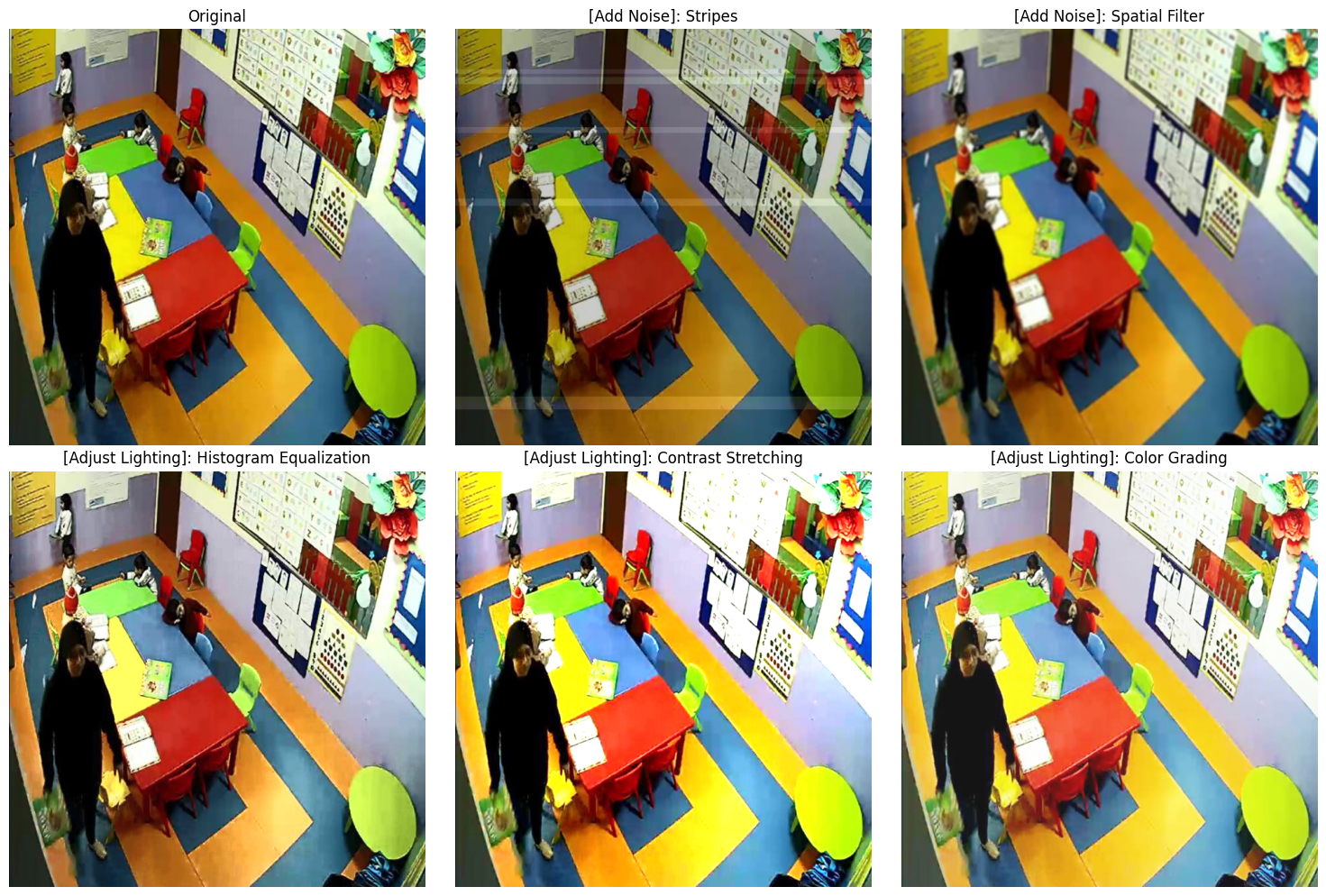}  %
  \caption{Examples of each effect used in Image-Level Degradations}
  \label{fig:add_noise_methods}
\end{figure}

\begin{figure}[!ht]
  \centering
  \begin{subfigure}{0.3\linewidth}
    \includegraphics[width=\linewidth]{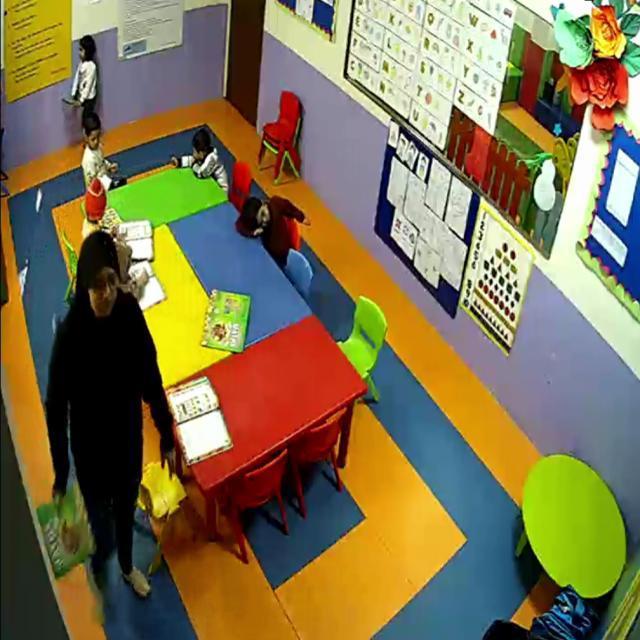}
    \caption{Original image from daycare training dataset without any modification}
  \end{subfigure}
  \begin{subfigure}{0.3\linewidth}
    \includegraphics[width=\linewidth]{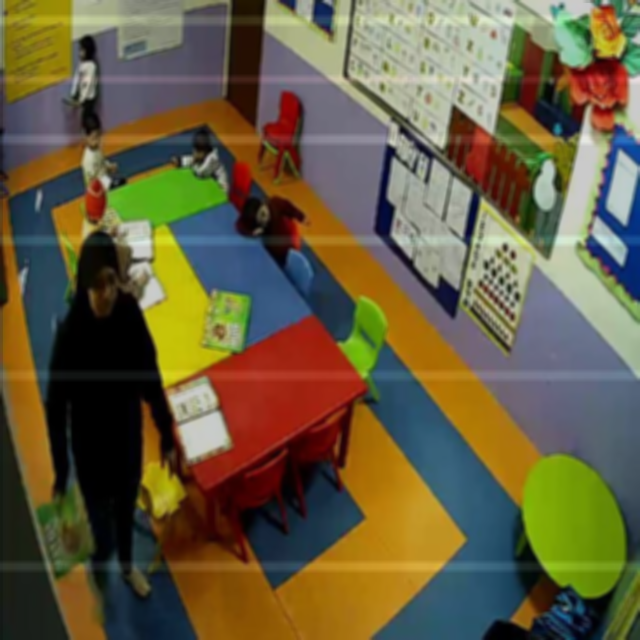}
    \caption{Noise injection using add strip and spatial filter to image}
  \end{subfigure}
  \begin{subfigure}{0.3\linewidth}
    \includegraphics[width=\linewidth]{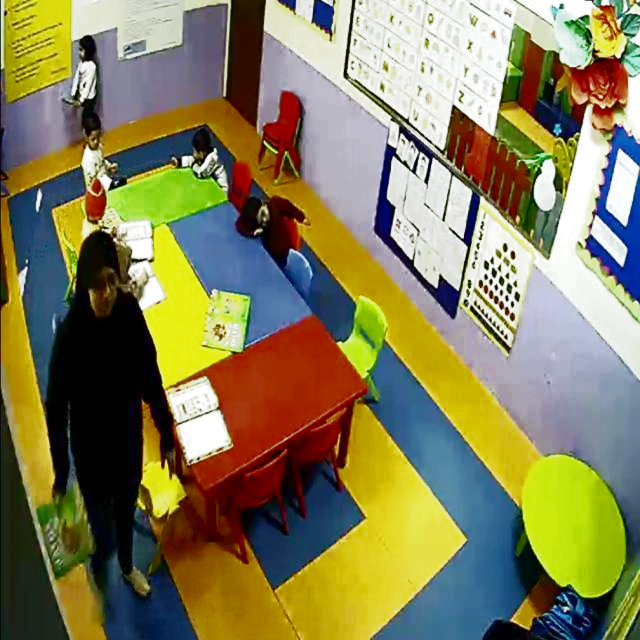}
    \caption{Light shift using color grading and histogram equalization}
  \end{subfigure}
  \caption{Image-level degradation examples.}
  \label{fig:sample_comparison}
\end{figure}

Examples of individual augmentation methods are presented in Figure~\ref{fig:add_noise_methods}, while the final augmented results incorporating image-level degradations are shown in Figure~\ref{fig:sample_comparison}. Together, these augmentations aim to enhance model robustness by exposing the detector to a wide range of difficult yet realistic visual scenarios during training. This augmentation pipeline is particularly vital for improving small object detection and generalization to unseen surveillance footage.


\subsection{Model Fine-Tuning}

To enable accurate and real-time child detection, we fine-tune YOLOv11n, a compact and efficient detector optimized for deployment on edge devices. The model provides a balanced compromise between inference speed and detection accuracy, suitable for resource-limited surveillance systems.

\paragraph{Training Setup}
We adopt transfer learning by initializing YOLOv11n with COCO-pretrained weights from Ultralytics to accelerate convergence and improve generalization. Training is performed using the official PyTorch implementation on the augmented dataset described in Section~\ref{subsection:DataAugmentation}, emphasizing challenging surveillance scenarios.

\subsection{SAHI Integration for Inference}

To enhance the detection of small or partially occluded children commonly observed in real-world surveillance settings, we incorporate Slicing Aided Hyper Inference (SAHI) during the inference phase. SAHI is a post-processing strategy that slices the input image into overlapping patches, performs inference on each slice, and merges results to recover missed detections.

\paragraph{Mechanism}
Let \( I \in \mathbb{R}^{H \times W \times 3} \) denote the input image. SAHI partitions \( I \) into \( n \) overlapping sub-images \( \{ I_i \}_{i=1}^n \), each of fixed size \( S \times S \), where:

\begin{equation}
S = \min(H, W) \cdot r, \quad \text{with } r \in (0, 1)
\end{equation}

An overlap ratio \( \gamma \in [0, 0.5] \) controls how much adjacent slices intersect, ensuring each region is seen under multiple receptive fields to aid detection at borders.

\paragraph{Inference and Merging}
Each sliced region \( I_i \) is independently processed by YOLOv11n. The resulting predictions are transformed back into the coordinate system of the original image. TTo remove redundant detections, Non-Maximum Suppression (NMS) is employed based on the Intersection over Union (IoU) criterion \( \tau \):

\begin{equation}
\text{IoU}(B_i, B_j) = \frac{|B_i \cap B_j|}{|B_i \cup B_j|} < \tau, \quad \forall i \neq j
\end{equation}

\paragraph{Advantages}
By focusing the detector on smaller regions, SAHI improves recall for small-scale or edge-truncated children that might otherwise be missed in full-frame inference. Although SAHI introduces additional computational overhead, it is justified in safety-critical applications like child surveillance, where missed detections could carry significant risk.

\section{Experiments and Evaluation}

\subsection{Experimental Setup}
To evaluate the effectiveness of the proposed augmentation approach and fine-tuned detector, we conduct a series of controlled experiments using the Roboflow Daycare dataset~\cite{roboflow_daycare}. The evaluation is performed on a held-out test set consisting of 564 surveillance-style images with 3755 annotated child instances. This test set is fixed across all experiments to ensure consistent comparability.

\subsection{Train Dataset Augmentation Strategy Configuration}

To evaluate the effect of different augmentation strategies, we conducted ablation experiments across three configurations: \textbf{Add Child-only Augmentation}, \textbf{Add Effect-only Augmentation}, and a combined \textbf{Add Child + Add Effect Augmentation}. Each configuration modifies the training set by introducing synthetic data according to the strategy defined in Section~\ref{subsection:DataAugmentation}.

Table~\ref{tab:augment_overview} summarizes the overall data statistics for each configuration. Dataset A introduces synthetic children into the training set, increasing the total image count by approximately 50\% (from 2289 to 3446) and object count by nearly 18\% (from 15518 to 18421). Dataset B applies only degradation effects such as noise and lighting shifts, producing a similar image count (3433) but without increasing the number of annotated objects. Dataset C combines both strategies, resulting in the largest dataset with 4590 images and 18421 labeled instances, providing the most comprehensive coverage of appearance and quality variations.

\begin{table}
\centering
\caption{Dataset Augmentation Experiment Overview}
\resizebox{\textwidth}{!}{
\begin{tabular}{|c|c|c|c|c|c|}
\hline
\textbf{Dataset} & 
\makecell{\textbf{Original} \\ \textbf{Images}} & 
\makecell{\textbf{Original} \\ \textbf{Objects}} & 
\makecell{\textbf{Augmentation Ratio} \\ \textbf{(train - overlap - edge - none - noise - light)}} & 
\makecell{\textbf{Final} \\ \textbf{Images}} & 
\makecell{\textbf{Final} \\ \textbf{Objects}} \\
\hline
\makecell{\textbf{A} \\ \textbf{(Child only)}} & 2289 & 15518 & 1 - 0.1 - 0.1 - 0.3 - 0 - 0 & 3446 & 18421 \\
\makecell{\textbf{B} \\ \textbf{(Effect only)}} & 2289 & 15518 & 1 - 0 - 0 - 0 - 0.25 - 0.25 & 3433 & 15518 \\
\makecell{\textbf{C} \\ \textbf{(Child + Effect)}} & 2289 & 15518 & 1 - 0.1 - 0.1 - 0.3 - 0.25 - 0.25 & 4590 & 18421 \\
\hline
\end{tabular}
}
\label{tab:augment_overview}
\end{table}

Table~\ref{tab:augment_breakdown} details the distribution of each augmentation component. Within the Add Child augmentation, three placement patterns are evenly distributed—228 images each for occlusion and edge truncation, and a larger subset of 686 centered placements, reflecting the higher frequency of central child appearances in the dataset. The corresponding object counts (407, 406, and 2090, respectively) indicate that centered placements tend to add more detectable regions per image. For Add Effect augmentation, both noise and light shifting are applied symmetrically, each contributing 572 augmented samples in Datasets B and C, ensuring a balanced distribution of environmental perturbations.
\begin{table}
\centering
\caption{Detailed Augmentation Experiment Breakdown}
\resizebox{\textwidth}{!}{
\begin{tabular}{|c|c|c|c|}
\hline
\textbf{Dataset} & 
\makecell{\textbf{Add Child - Images} \\ \textbf{(Overlap / Edge / Center)}} & 
\makecell{\textbf{Add Child - Objects} \\ \textbf{(Overlap / Edge / Center)}} & 
\makecell{\textbf{Add Effect - Images} \\ \textbf{(Noise / Light Shift)}} \\
\hline
\textbf{A} & 228 / 228 / 686 & 407 / 406 / 2090 & 0 / 0 \\
\textbf{B} & 0 / 0 / 0 & 0 / 0 / 0 & 572 / 572 \\
\textbf{C} & 228 / 228 / 686 & 407 / 406 / 2090 & 572 / 572 \\
\hline
\end{tabular}
}
\label{tab:augment_breakdown}
\end{table}
\FloatBarrier

Overall, Dataset C yields the most diverse and balanced composition, combining object-level variations (synthetic child placements) with environmental distortions (lighting and noise). This comprehensive augmentation is expected to produce a model more robust to complex real-world surveillance conditions.

\subsection{Model Evaluation}
\label{subsection:ModelEvaluation}

To assess the performance of the trained models, we adopt standard object detection metrics: \textbf{Precision}, \textbf{Recall}, \textbf{mAP@0.5}, and \textbf{mAP@0.5:0.95}. All models are trained for \textbf{100 epochs} with a \textbf{batch size: 16}. We evaluate five different training configurations. Table~\ref{tab:evaluation_results} presents the quantitative comparison across all setups.

\begin{enumerate}
  \item \textbf{YOLOv11n (baseline)}: trained on the original unaugmented dataset.
  \item \textbf{YOLOv11n + Dataset A}: trained on data augmented with synthetic children only.
  \item \textbf{YOLOv11n + Dataset B}: trained on data augmented with image-level noise and light effects only.
  \item \textbf{YOLOv11n + Dataset C}: trained on combined synthetic children and image-level effects.
  \item \textbf{YOLOv11n + Dataset C + SAHI}: the best trained model above, enhanced with SAHI inference to better detect small or truncated children.
\end{enumerate}

The evaluation results in Table~\ref{tab:evaluation_results} show consistent improvements over the baseline when applying targeted data augmentations. Both Dataset A (synthetic child compositing) and Dataset B (photometric degradation) individually yield modest gains, particularly in recall and mAP@0.5:0.95. Dataset A enhances spatial diversity and helps the model learn to detect occluded or variably positioned children, while Dataset B improves robustness to lighting and visual distortions.

\begin{table}
\centering
\tiny
\caption{Model Evaluation on Surveillance-Style Daycare Test Set}
\label{tab:evaluation_results}
\resizebox{\linewidth}{!}{
\begin{tabular}{|l|c|c|c|c|}
\hline
\textbf{Model} & 
\textbf{Precision} & 
\textbf{Recall} & 
\makecell{\textbf{mAP} \\ \textbf{@0.5}} &
\makecell{\textbf{mAP} \\ \textbf{@0.5:0.95}} \\
\hline
YOLOv11n (Baseline - Original Dataset) & \textbf{0.947} & 0.925 & 0.963 & 0.760 \\
YOLOv11n + Dataset A & 0.942 & \textbf{0.937} & 0.964 & 0.764 \\
YOLOv11n + Dataset B & 0.939 & 0.935 & 0.961 & 0.773 \\
YOLOv11n + Dataset C & 0.943 & 0.934 & 0.964 & 0.779 \\
YOLOv11n + Dataset C + SAHI & 0.946 & 0.933 & \textbf{0.967} & \textbf{0.783} \\
\hline
\end{tabular}
}
\end{table}

\begin{figure*}[ht]
  \centering
  \begin{subfigure}{0.45\linewidth}
    \centering
    \includegraphics[width=\linewidth]{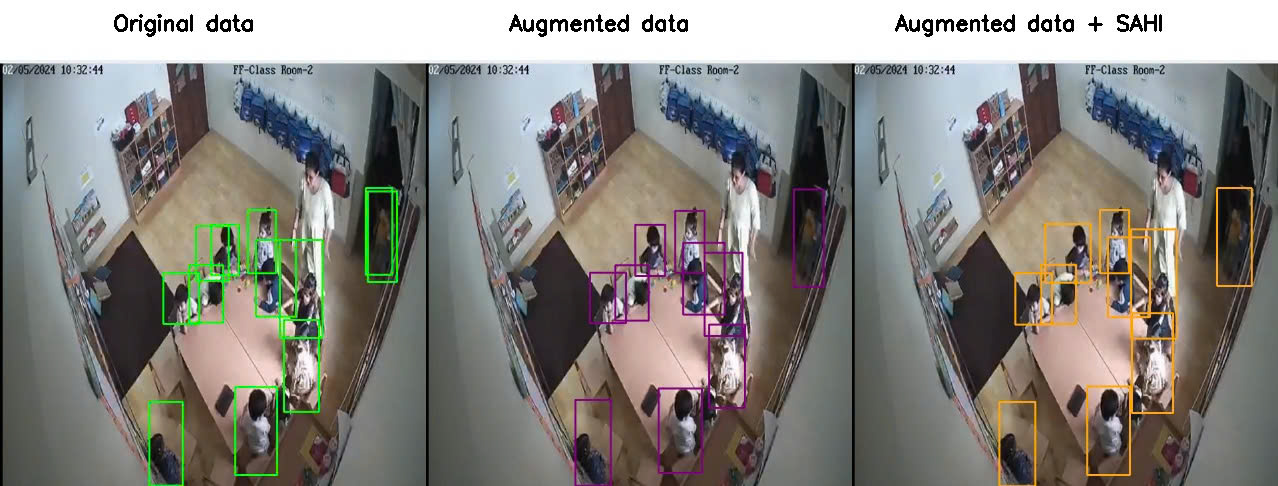}
    \caption{YOLOv11n output on surveillance image A.}
  \end{subfigure}
  \hfill
  \begin{subfigure}{0.45\linewidth}
    \centering
    \includegraphics[width=\linewidth]{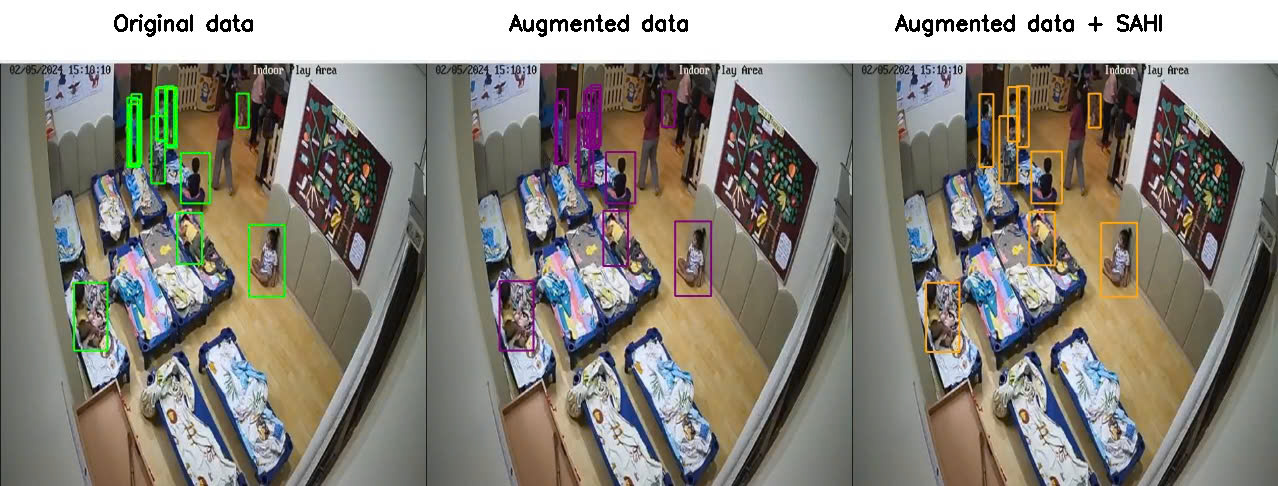}
    \caption{YOLOv11n + SAHI output on surveillance image B.}
  \end{subfigure}
  \caption{Qualitative detection results from fine-tuned YOLOv11n models.}
  \label{fig:qualitative_examples}
\end{figure*}

The highest overall performance is achieved with Dataset C, which combines both augmentation strategies, demonstrating their complementary effects. Precision is maintained while recall and mAP metrics improve, especially at higher IoU thresholds. Finally, integrating SAHI further boosts mAP@0.5:0.95 by improving detection of small and edge-truncated children, critical for daycare surveillance settings, without significantly sacrificing computational efficiency. Examples of output from the fine-tuned YOLOv11n models are shown in Fig.~\ref{fig:qualitative_examples}.

\section{Conclusion and Future Work}

This study presents a data augmentation pipeline designed to improve child detection in surveillance settings using the lightweight YOLOv11n model. By compositing synthetic child instances into real daycare scenes and applying photometric degradations, the approach mitigates common challenges such as occlusion, motion blur, and poor lighting. Experiments show consistent yet modest performance gains, primarily limited by the single-camera nature of the Roboflow Daycare dataset, which restricts background diversity and cross-view generalization. Nonetheless, combining spatial and photometric augmentations with SAHI inference improves robustness to small and partially visible children.

Future work will extend this study through multi-camera data collection, larger and more diverse datasets, and advanced fine-tuning approaches like domain-adaptive training or self-supervised pretraining. Overall, the proposed framework offers a practical and extensible solution for child detection in safety-critical surveillance systems, emphasizing the value of diverse and realistic augmentation for robust performance.

%
%
%
%

\end{document}